\documentclass[11pt]{article}

\usepackage[preprint]{acl}

\usepackage{times}
\usepackage{latexsym}

\usepackage[T1]{fontenc}
\usepackage[utf8]{inputenc}

\usepackage{microtype}

\usepackage{inconsolata}

\usepackage{graphicx}

\usepackage{booktabs} % for professional tables
\usepackage{multirow}
\usepackage{bm}
\usepackage{xcolor}
\usepackage{amsmath}
\usepackage{amsthm}
\usepackage{subcaption}

\usepackage{caption}
\usepackage{float} 
\usepackage{subcaption}
\usepackage{amsmath,amssymb,amsfonts}
\usepackage{algorithmic}
\usepackage{textcomp}
\usepackage{bm}
\usepackage{amsfonts}
\usepackage[table]{xcolor}
\usepackage{diagbox}
\usepackage{booktabs}
\usepackage[normalem]{ulem}
\usepackage{multirow}
\usepackage{pifont}
\usepackage{enumitem}
\usepackage{booktabs}
\usepackage{amssymb}
\usepackage{subcaption}
\usepackage{wrapfig}
\usepackage{tikz}
\theoremstyle{plain}
\theoremstyle{definition}
\theoremstyle{remark}
\usepackage{amsmath}
\usepackage{amssymb}
\usepackage{mathtools}
\usepackage{amsthm}
\usepackage{colortbl}
\usepackage{tikz}      % 用于绘制图形
\usepackage{colortbl}
\usepackage{float}
\usepackage{wrapfig}
\usepackage{hyperref}
\usepackage{listings}
\usepackage{tcolorbox}
\usepackage{pifont}
\usepackage{multicol}

\usepackage{subcaption}
\newcommand{\circnum}[1]{\raisebox{-0.3ex}{\scalebox{1.3}{\ding{#1}}}}

\title{\texttt{Q-realign}: Piggybacking Realignment on Quantization for Safe and Efficient LLM Deployment}

\author{
 Qitao Tan\textsuperscript{1} \quad
Xiaoying Song\textsuperscript{2} \quad
Ningxi Cheng\textsuperscript{1} \quad
\textbf{Ninghao Liu}\textsuperscript{3} \quad \\
\textbf{Xiaoming Zhai}\textsuperscript{1} \quad
\textbf{Lingzi Hong}\textsuperscript{2} \quad
\textbf{Yanzhi Wang}\textsuperscript{4} \quad
\textbf{Zhen Xiang}\textsuperscript{1} \quad
\textbf{Geng Yuan}\textsuperscript{1} \\
\textsuperscript{1}University of Georgia \quad
\textsuperscript{2}University of North Texas \\
\textsuperscript{3}Hong Kong Polytechnic University \quad
\textsuperscript{4}Northeastern University \\[-2.0em]
}

\begin{document}
\maketitle
\begin{abstract}
% Public large language models (LLMs) typically undergo safety alignment during pretraining to promote compliant and responsible behavior prior to downstream adaptation. However, practical deployment often requires task-specific fine-tuning, which has been shown to substantially compromise the original alignment and introduce safety risks at deployment time.
% Most existing defense approaches either integrate safety recovery directly into the fine-tuning pipeline or rely on priors derived from fine-tuning for post-processing. As a result, safety recovery remains tightly coupled with training, leading to high computational costs and complex workflows. 
Public large language models (LLMs) are typically safety-aligned during pretraining, yet task-specific fine-tuning required for deployment often erodes this alignment and introduces safety risks. Existing defenses either embed safety recovery into fine-tuning or rely on fine-tuning-derived priors for post-hoc correction, leaving safety recovery tightly coupled with training and incurring high computational overhead and a complex workflow. To address these challenges, we propose \texttt{Q-realign}, a post-hoc defense method based on post-training quantization, guided by an analysis of representational structure. By reframing quantization as a dual-objective procedure for compression and safety, \texttt{Q-realign} decouples safety alignment from fine-tuning and naturally piggybacks into modern deployment pipelines. Experiments across multiple models and datasets demonstrate that our method substantially reduces unsafe behaviors while preserving task performance, with significant reductions in memory usage and GPU hours. Notably, our approach can recover the safety alignment of a fine-tuned 7B LLM on a single RTX 4090 within 40 minutes. Overall, our work provides a practical, turnkey solution for safety-aware deployment 
\footnote{Our code is released at \url{https://github.com/Skilteee/Q-Realign}.}.
\end{abstract}

\section{Introduction}

\begin{figure*}[t]
\centering
\vspace{-20pt}
\includegraphics[width=0.8\linewidth]{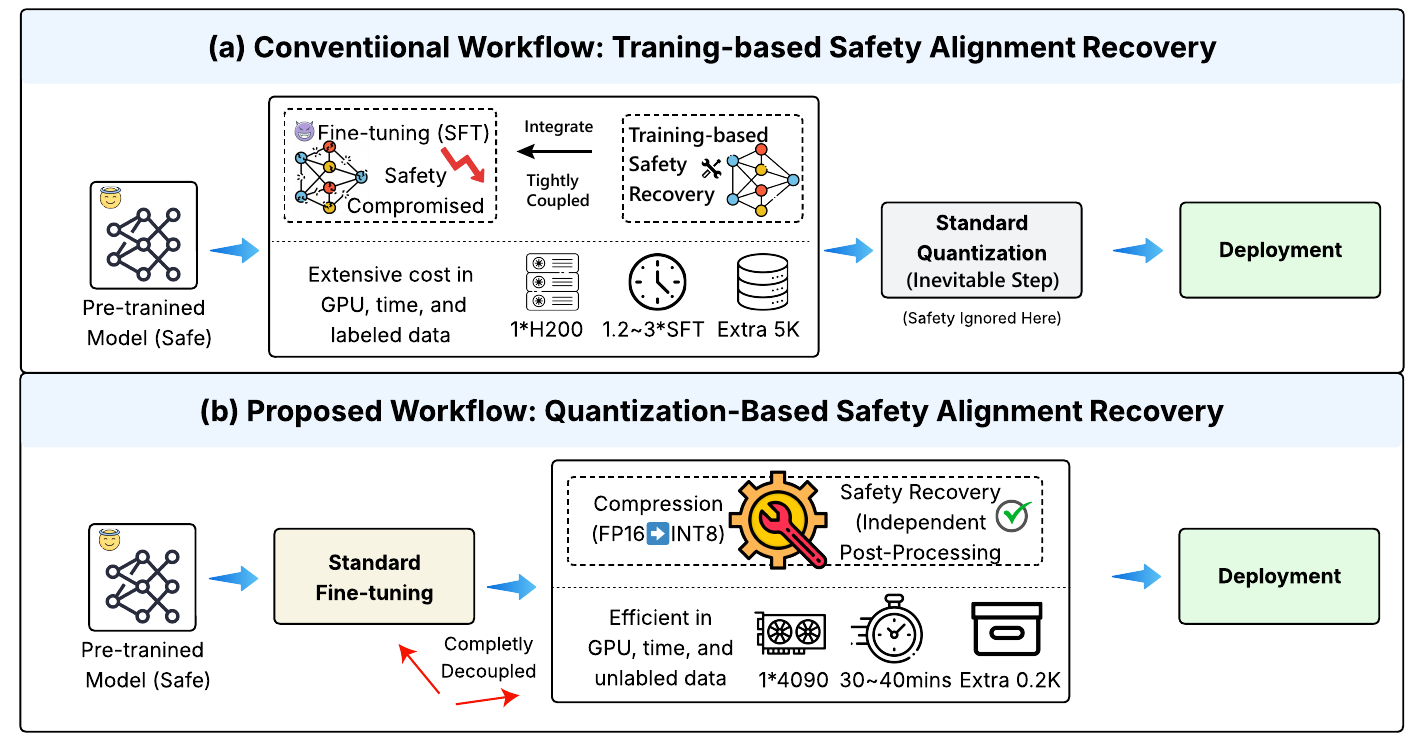}
\caption{
Comparison between conventional training-based safety recovery and our quantization-based approach. By piggybacking alignment recovery onto the standard post-training quantization step, our method fully decouples safety recovery from fine-tuning, eliminating the need to re-enter the fine-tuning process. This design substantially reduces computational overhead and simplifies the deployment workflow.
}
% \vspace{-10pt}
\label{zo_fo}
\end{figure*}

%  这一段需要先强调fine-tuning过程在研究和商业中都应用广泛，然后说fine-tuning会导致safety risk，这很容易被malicous用户利用，还会引起benign用户的concern，所以解决fine-tuning中的安全问题是重要的。
Fine-tuning of Large Language Models (LLMs) has become a critical way for a wide range of today's specialized downstream applications. To ensure that LLMs can generate helpful content and refuse malicious requests (e.g., how to make a bomb), modern LLMs typically require alignment training, usually by RLHF~\citep{ouyang2022training}. However, recent studies point out a safety issue~\citep{qi2023fine, yang2023shadow, huang2024harmful, li2024peft, ye2024emerging}: simply fine-tuning aligned LLMs on a dataset that contains a few malicious samples, or even on a pure benign dataset, can significantly compromise the alignment and cause a noticeable rise in unsafe behaviors. Moreover, as datasets may contain latent unsafe data that is difficult for users to detect~\citep{wang2025panacea}, even benign model creators can encounter safety degradation issues after fine-tuning.

Existing solutions addressing safety degradation caused by fine-tuning on aligned models can be broadly categorized according to the stage at which they operate~\citep{huang2024vaccine}. The first category intervenes during fine-tuning, directly modifies the training procedure, formulates training as a multi-objective optimization problem that jointly balances downstream task performance and safety alignment~\citep{mukhoti2023fine,zong2024safety,lyu2024keeping,du2024towards,qi2024safety,li2025salora, yi2025gradient}. 
The second category applies post-hoc defenses after fine-tuning, but typically depends on fine-tuning-derived signals and requires additional auxiliary training or adaptation, preventing truly plug-and-play safety recovery~\citep{du2024mogu,bhardwaj2024language,yi2024safety, wang2025panacea}. 
Though studies are exploring one-shot post-hoc defenses~\citep{hsu2024safe,huangantidote, yi2025nlsr}, their empirical performance is generally unfavorable under realistic attack scenarios.

Despite operating at different stages, methods in both categories remain tightly coupled with fine-tuning. From a computational perspective, this tight coupling leads to substantial computational overhead, including additional training iterations, increased memory consumption, and extended GPU usage. From a workflow perspective, existing defenses are therefore not plug-and-play. During-fine-tuning methods require retraining by construction, while post-hoc approaches typically depend on fine-tuning-derived signals or additional adaptation steps, which likewise necessitate re-entering the fine-tuning pipeline. For example, when safety vulnerabilities are identified after deployment, existing defenses often require re-conducting fine-tuning, rather than enabling lightweight, modular safety recovery. Motivated by these limitations, we ask \emph{whether there exists a method to effectively mitigate safety degradation caused by fine-tuning, while decoupling from fine-tuning for better computational and workflow efficiency?}

Driven by this observation, we seek a safety recovery method that \textbf{operates at an inherent post-training stage between fine-tuning and deployment}, and avoids looping back to the fine-tuning stage.
Notably, model quantization is an almost inevitable step in modern LLM deployment pipelines, yet it has been largely overlooked from a safety recovery perspective. This motivates us to revisit quantization as a dual-purpose operation that not only facilitates model compression for efficient deployment but also provides an opportunity for alignment recovery. In particular, we leverage post-training quantization (PTQ), a de facto standard in modern LLM deployment, to naturally support safety recovery. The workflow of our method is shown in Figure~\ref{zo_fo}.
Compared to fine-tuning, PTQ incurs substantially lower computational cost and does not introduce additional training stages. As a result, our method achieves alignment recovery while preserving the original role of quantization, leading to significant gains in both computational and workflow efficiency.

To understand how quantization can recover safety alignment after fine-tuning, we analyze the representational structure of LLMs before and after fine-tuning, revealing how fine-tuning compromises the semantic separability underlying alignment. 
% Our layer-wise activation analysis reveals that, in aligned models, benign and malicious inputs exhibit marked separability in intermediate activations, whereas fine-tuning tends to erode this structure. 
Motivated by this observation, we introduce learnable layer-wise quantization parameters to selectively constrain activations and recover this separability, thereby recovering alignment. Comprehensive experimental results show that our method significantly reduces unsafe behaviors while preserving task performance, with significant reductions in memory usage and GPU hours. In practice, our approach recovers the safety alignment of a fine-tuned 7B LLM on a single RTX 4090 within 40 minutes.

We summarize our contributions as follows: 1) We introduce an analysis to uncover how fine-tuning compromises the safety alignment from the perspective of intermediate activation. 2) We propose \texttt{Q-realign}, a PTQ-based defense framework for fine-tuning attacks, which can naturally be integrated into the deployment pipeline.
3) \texttt{Q-realign} consistently exceeds existing baselines in safety alignment recovery while maintaining fine-tuning accuracy, with a significant memory and GPU hours improvement. These advantages hold across diverse tasks and LLM architectures.

\section{Preliminaries and Activation Analysis}

% \subsection{Inefficiency of Existing Methods}

% To mitigate safety degradation caused by fine-tuning, for defense, assume that there is an alignment task and a fine-tuning task. For existing methods, 

% 只要涉及到alignment的变化，就要reconduct the whole fine-tune process，举几个case的例子，强调除了alignment之外，还需要执行ft的cost. 有个alignment task和ft task，现有的方法和ft coupled，alignment task不能单独执行，对于如果一个model需要额外的alignment task，我们的就可以alignment task only. 然后举几个例子

\subsection{Threat Model and Assumptions}

Our threat model and experimental setup are defined as follows. A model provider fine-tunes a pre-aligned LLM (e.g., LLaMA2-7B-chat) on task-specific data and deploys it as an online service accessed via an API. In this setting, the provider remains responsible for harmful outputs generated after fine-tuning and may face significant governance and legal risks~\citep{reuel2024open}.

Unlike prior work that assumes access to large labeled alignment datasets with harmful instructions paired with safe responses~\citep{hsu2024safe, huang2024lazy, wang2025panacea}, our method operates under a weaker assumption: it requires only harmful instructions and substantially less alignment data. This makes our approach more practical for real-world deployment, where collecting high-quality safety annotations is costly. 
We present our detailed related work section in Appendix~\ref{related_work}.
% We give some examples of data in Appendix~\todo{XXX}

\subsection{How Fine-tuning Compromises Safety Alignment?}
\label{analysis}

In this section, we investigate how fine-tuning compromises the safety alignment of LLMs by examining changes in their representational structure. Specifically, we analyze the separability of layer-wise intermediate activations of benign and malicious inputs before and after fine-tuning. The separability is measured from two complementary perspectives: \emph{spatial separability}, capturing geometric separation in representation space, and \emph{semantic separability}, reflecting semantically separable information.

\textbf{Analysis Method:}   Spatial separability is measured by Fisher's linear discriminant ratio~\citep{fisher1936use}. Given the token-averaged $l$-th layers' activation $\mathbf{h}^{(l)} \in \mathbb{R}^{d}$, the spatial separability is formulated as:
\begin{equation}
\small
\mathcal{S}^{(l)}
=
\frac{
\left\|
\mathbb{E}_{\mathbf{h} \sim \mathcal{H}^{(l)}_{\text{benign}}}[\mathbf{h}]
-
\mathbb{E}_{\mathbf{h} \sim \mathcal{H}^{(l)}_{\text{malicious}}}[\mathbf{h}]
\right\|_2
}{
\mathbb{E}_{\mathbf{h} \sim \mathcal{H}^{(l)}}
\left[
\left\|
\mathbf{h}
-
\boldsymbol{\mu}^{(l)}_{y(\mathbf{h})}
\right\|_2
\right]
+
\varepsilon
}
\end{equation}
where $\mathcal{H}^{(l)}_{\text{benign}}$ and $\mathcal{H}^{(l)}_{\text{malicious}}$ denote the benign and malicious activation sets, respectively, $\boldsymbol{\mu}^{(l)}_{y(\mathbf{h})}$ is the centroid of $\mathbf{h}$’s class, and $\varepsilon$ is a small constant for numerical stability.

A higher $\mathcal{S}^{(l)}$ indicates that benign and malicious representations are more compact within each class and more distant across classes, reflecting stronger separability.

We measure semantic separability using the refusal rate, which reflects how often malicious inputs are mapped to semantic regions associated with refusal behavior. For each malicious input, we perform layer-wise decoding from the intermediate activations at layer $l$ and check whether the generated output contains explicit refusal expressions. The refusal rate at layer $l$ is defined as the proportion of malicious inputs whose decoded outputs exhibit refusal-related language. Intuitively, if intermediate activations encode clear semantic separation between benign and malicious inputs, this separation becomes behaviorally explicit during decoding, resulting in a higher refusal rate. Refusal detection is implemented using a keyword-based method following \citet{chao2023jailbreaking}.

\begin{figure}[t]
    \centering
    % \hspace{-25pt}
	\begin{subfigure}{0.49\linewidth}
		\centering
		\includegraphics[width=1.0\linewidth]{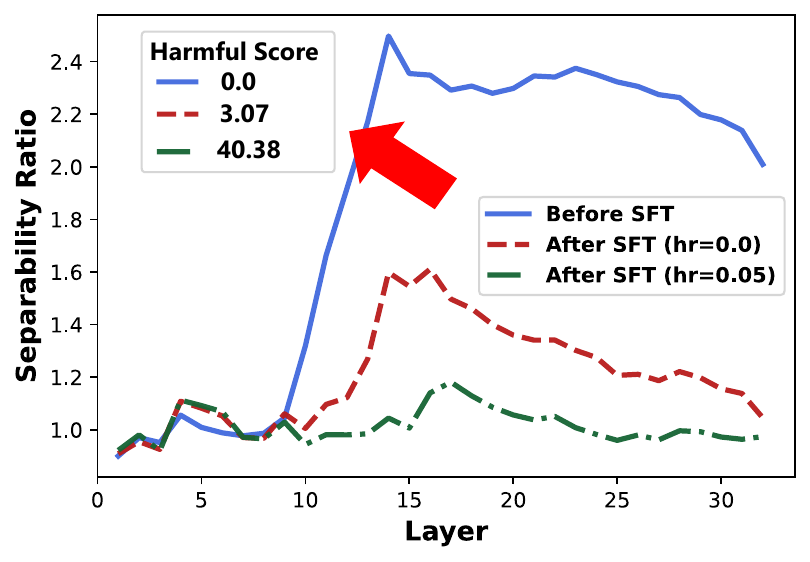}
		\caption{}
        \label{analysis_a}
	\end{subfigure}
    % \hspace{15pt}
	\begin{subfigure}{0.49\linewidth}
		\centering
		\includegraphics[width=1.0\linewidth]{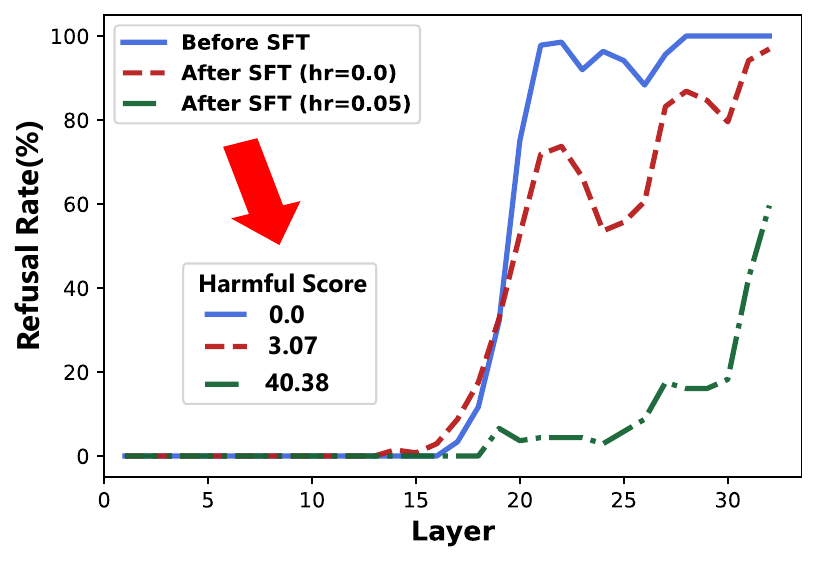}
        \caption{}
		% \caption{On-the-fly generation reuse strategy}
        \label{analysis_b}
	\end{subfigure}
    
    \caption{
    Layer-wise separability and refusal rate across pre-trained and fine-tuned models with varying harmful ratios (hr). Lower harmful score indicates a safer model.
    }
    \vspace{-15pt}
\end{figure}

\textbf{Analysis Results:} Figure~\ref{analysis_a} and Figure~\ref{analysis_b} illustrate the spatial separability and semantic separability before and after fine-tuning, respectively.
As shown in Figure~\ref{analysis_a}, the aligned pre-trained model exhibits higher separability in the middle and deeper layers, whereas fine-tuning substantially reduces separability across most layers, with a more pronounced decline when a higher fraction of malicious data is used (e.g., hr = 0.05). This reduction in spatial separability is consistent with the observed increase in the harmful score. These trends suggest that fine-tuning blurs the separation between benign and malicious behaviors, undermining safety alignment and increasing vulnerability to harmful inputs (refer to Figure~\ref{analysis_activation} and Figure~\ref{Analysis} for 2D visualization).

% In Figure~\ref{analysis_a}, we measure layer-wise spatial separability. The aligned pre-trained model exhibits consistently higher separability across layers, indicating that intermediate activations induced by benign and malicious inputs are well separated. After fine-tuning, this separability ratio decreases substantially, with a more pronounced degradation when fine-tuning includes 5\% malicious samples. The reduced layer-wise separability suggests that fine-tuning blurs the representational distinction between benign and malicious behaviors, which correlates with a higher attack success rate.

In Figure~\ref{analysis_b}, we report the layer-wise refusal rate as a behavioral measure of semantic separability. In the aligned pre-trained model, decoding directly from intermediate activations, rather than only from the final layer, already yields a very high refusal rate in the middle and upper layers (e.g., layer 20+). Specifically, activations from these layers lead to refusal of nearly all malicious inputs, indicating that malicious activations are mapped to semantically distinct regions associated with refusal behavior. Although decoding from intermediate layers may produce less fluent outputs due to bypassing later decoding stages, the model consistently exhibits refusal behavior, suggesting that semantic separability is already established at these layers.

In contrast, this early-layer refusal capability deteriorates rapidly after fine-tuning. Fine-tuned models exhibit substantially lower refusal rates when decoding from intermediate layers, indicating that fine-tuning weakens the semantic separability previously encoded in these representations. Notably, when fine-tuning includes a higher fraction of malicious data, the model fails to reliably refuse malicious inputs even when decoding from later layers (e.g., layer 30), with refusal rates remaining below 20\%. This pronounced decline suggests that fine-tuning increasingly entangles the semantic representations of benign and malicious inputs, undermining the model’s ability to trigger refusal behavior.

% In contrast, this early rejection capability rapidly diminishes after fine-tuning. Fine-tuned models show substantially lower rejection rates when decoding from intermediate layers, suggesting that fine-tuning erodes the semantic separability encoded in these representations. Notably, after fine-tuning with 5\% malicious samples, the model fails to reliably reject malicious inputs even when decoding from later layers (e.g., layer 30), with rejection rates remaining below 20\%. These results indicate that fine-tuning compresses or entangles malicious and benign.

% In Figure~\ref{analysis_b}, we report the layer-wise rejection rate as a metric of semantic separability. In the aligned pre-trained model, decoding directly from intermediate activations-rather than only from the final layer-already yields a high rejection rate. For example, activations at middle layers (e.g., layer 23) lead to rejection of nearly all malicious inputs, indicating that malicious representations are already semantically well separated at these layers. Although decoding from intermediate layers may result in less fluent outputs due to bypassing later decoding stages, the model consistently exhibits refusal behavior (see Appendix \todo{XXX} for examples).

\textbf{Takeaway:} From Figure~\ref{analysis_a} and Figure~\ref{analysis_b}, we draw the following observations.
\circnum{172} In aligned LLMs, intermediate activations induced by benign and malicious inputs exhibit high layer-wise spatial separability.
\circnum{173} This spatial separability further manifests as semantic separability: decoding directly from intermediate layers already yields strong refusal behavior, indicating that malicious inputs are mapped to distinct semantic regions early in the network.
\circnum{174} Fine-tuning substantially degrades both properties. After fine-tuning, the layer-wise spatial and semantic separability between benign and malicious activations is markedly reduced, accompanied by a sharp increase in harmful score.

Motivated by these findings, we conjecture that safety degradation after fine-tuning stems from the loss of separability in intermediate activations. While semantic separability directly reflects safety alignment but is difficult to manipulate, spatial separability offers a more tractable structural proxy. This leads to the question:
\emph{Can restoring spatial separability in intermediate activations restore semantic separability, and thereby indirectly recover safety alignment?}

% Motivated by these findings, we conjecture that safety degradation after fine-tuning is closely tied to the loss of intermediate activation separability. While semantic separability is likely a direct manifestation of alignment, it isn't easy to explicitly control. This naturally leads to the following question:
% \emph{Can we recover spatial separability of intermediate activations to recover semantic separability, and thereby indirectly recover safety alignment?}

\section{Methodology}

\subsection{Representation and Extraction of Spatial Separability}

Based on our analysis of intermediate activations in aligned pre-trained LLMs, we observe that activations induced by benign and malicious inputs exhibit clear layer-wise spatial separability, which is largely linear in the activation space. To explicitly characterize and extract this separability, we adopt a Sparse Logistic Regression (SLR) model, following prior work on probing toxic representations~\cite{hu2024toxicity}.

Given the token-averaged activation at layer $l$, denoted as $\mathbf{h}^{(l)} \in \mathbb{R}^{d}$, the corresponding SLR decision function is defined as
\begin{equation}
\small
    \mathrm{SLR}(\mathbf{h}^{(l)}) = \mathbf{w}^{(l)\top} \mathbf{h}^{(l)} + b^{(l)},
\end{equation}
where $\mathbf{w}^{(l)} \in \mathbb{R}^{d}$ and $b^{(l)} \in \mathbb{R}$ denote the weight vector and bias term of the SLR model at layer $l$, respectively. 

Importantly, our goal is not to use SLR as a downstream classifier, but rather to leverage the learned hyperplane as a geometric abstraction of spatial separability in the activation space. This abstraction provides a structured representation and serves as a foundation for designing alignment recovery quantization strategies.

In practice, we train a lightweight SLR probe for each layer on an aligned pre-trained model and reuse these SLR to guide alignment recovery on fine-tuned models. Since SLRs are linear models trained on frozen calibration activations without updating LLM parameters, their training incurs negligible overhead compared to fine-tuning (i.e., about ten minutes on CPU for a 7B model). Moreover, the learned SLRs are specific to a pre-trained aligned LLM, and are reusable for a model fine-tuned on different downstream datasets.

\subsection{Recovering LLM Alignment via Post-training Quantization}

% we use post-training quantization for its high efficiency, as it avoids retraining of the model. Specifically, we use the quantization schema in OmniQuant~\citep{shao2023omniquant}, which defines a block-wise reconstruction loss is formulated as:

We revisit quantization as a dual-purpose operation that not only facilitates model compression for efficient deployment but also provides an opportunity for alignment recovery. As a proof of concept, we adopt the efficient PTQ scheme from OmniQuant~\citep{shao2023omniquant}. While we instantiate our approach within the OmniQuant framework, the proposed method can also be incorporated into other quantization schemes with learnable quantization parameters.

To preserve model accuracy after quantization, we follow prior work and formulate a block-wise reconstruction loss:
% \vspace{-10pt}
\begin{equation}
\label{eq3}
\small
% \arg\min_{s^{(l)},\Theta^{(l)}}
% \mathcal{L}_{1} = 
||
\mathcal{F}(W^{(l)}, \mathbf{h}^{(l)}) - \mathcal{F}\big(Q_w(W^{(l)}; s^{(l)}, \gamma^{(l)}),\, Q_a(\mathbf{h}^{(l)}; s^{(l)})\big)
||,
\end{equation}
where $\mathcal{F}$ represents the mapping function for a transformer block in the LLM, $W$ and $X$ are full-precision model weight and activation, $Q_w(\cdot)$ and $Q_a(\cdot)$ represent weight and activation quantizer, respectively, $s^{(l)}$ is the learnable quantization smoothing scalar for equivalent transformation~\citep{xiao2023smoothquant}, and $\gamma^{(l)}$ is learnable quantization parameter for weight clipping. More details related to quantization are presented in Appendix~\ref{quantization}.

Moreover, to re-separate the activations of benign and malicious inputs after quantization, we introduce an additional alignment recovery objective. Specifically, we employ the Softplus loss~\citep{dugas2000incorporating} to push quantized activations away from the decision boundary learned by the Sparse Logistic Regression (SLR) probe. Formally, the loss is defined as
\begin{equation}
\label{eq4}
\small
% \mathcal{L}_{2}=
\log\!\left(1+\exp\!\left(-(\mathbf{w}^{(l)\top}\hat{\mathbf{h}}^{(l)}+b^{(l)})\right)\right),
\end{equation}
where $\mathbf{w}^{(l)}$ and $b^{(l)}$ is the parameter of the SLR of the $l$-th layer, and $\hat{\mathbf{h}}^{(l)}$ is the quantized activation of the $l$-th layer.

For each layer, quantization is guided by two objectives. The first is to preserve model accuracy by reconstructing activations after quantization, and the second is to re-separate the activations of benign and malicious inputs to recover safety alignment. These two objectives are inherently in conflict: preserving accuracy encourages activations to remain close to their original values, whereas re-separation requires actively shifting activations to restore alignment-related activation structure. 

To resolve this conflict, we build on the observation that model helpfulness is primarily determined by benign inputs, while safety violations are mainly associated with malicious inputs.
Accordingly, we adopt a class-conditional strategy: we reconstruct the activations of benign inputs to preserve task performance, while pushing the activations of malicious inputs away from benign regions in the representation space to recover spatial separability and safety alignment. This leads to the following optimization objective:
\begin{equation}
\small
\begin{aligned}
\arg\min_{\Theta^{(l)}}
\;&\mathbb{E}_{(\mathbf{h}^{(l)}, y)}
\Big[
\underbrace{
(1-y)\,\lVert W^{(l)} \mathbf{h}^{(l)} - \hat{W}^{(l)} \hat{\mathbf{h}}^{(l)} \rVert_2^2
}_{\text{reconstruction}}
\\
&\quad +\;
\underbrace{
y\,\lambda \log\!\big(1 + \exp\!\big(-(\mathbf{w}^{(l)\top}\hat{\mathbf{h}}^{(l)}+b^{(l)}))\big)\big)
}_{\text{re-separation}}
\Big]
\end{aligned}
\end{equation}
where $\hat{W}^{(l)}$ and $\hat{\mathbf{h}}^{(l)}$ are quantized weight and activation, obtained by learnable quantization parameters $\Theta^{(l)}=\{s^{(l)}, \gamma^{(l)}\}$, the binary label $y=\{0, 1\}$ indicates the activation type, with $y=1$ corresponding to malicious samples.

\section{Experiments}

We present a comprehensive evaluation of \texttt{Q-realign}, reporting results on different harmful ratios, fine-tuning datasets, and LLMs. (Sections~\ref{main_result}), followed by an efficiency analysis including memory and speed (Section~\ref{overhead}). We also provide different ablation studies to assess generalizability (Appendix~\ref{ablation}). Moreover, we provide a statistical analysis and visualization to illustrate the effectiveness of our method (Section~\ref{more_analysis}).

\subsection{Experimental Setup}
\textbf{Models and datasets.} Following prior work~\citep{lyu2024keeping}, we use the aligned pre-trained LLaMA2-7B-Chat~\citep{touvron2023llama} as the primary base model for fine-tuning. To assess the generalizability of our approach, we additionally conduct experiments on other widely used LLMs, including Gemma2-9B-it~\citep{team2024gemma} and Qwen2.5-7B-it~\citep{yang2025qwen2}. For datasets, we use Alpaca~\citep{alpaca} as the default fine-tuning dataset, and further evaluate on SST-2~\citep{socher2013recursive} and GSM8K~\citep{cobbe2021gsm8k}, following the experimental protocols of~\citep{rosati2024representation, huang2024lazy, wang2025panacea}. Models fine-tuned on Alpaca are evaluated using the MMLU benchmark~\citep{hendrycks2020measuring} following~\citet{pan2024lisa}, while models fine-tuned on SST-2 and GSM8K are evaluated on their corresponding test sets.

\textbf{Implementation details and Evaluation metrics.} We evaluate under the scenario following~\citep{rosati2024representation, huang2024lazy, lyu2024keeping}, a user or service provider fine-tunes on a dataset that contains a portion of malicious samples, and the portion named harmful ratio (hr). We use the harmful score (hs) to evaluate the safety of a model, and we test safety on the subset of the AdvBench benchmark curated by~\citet{zou2023universal}. We use LoRA~\citep{hu2022lora} for efficient fine-tuning and then quantize the model to W8A8 (i.e., 8-bit weight and 8-bit activation) by default and maintain other methods in floating point. All our results are the average of three runs on A6000 GPUs by default. More details, including baselines, are included in Appendix~\ref{settings}.

\subsection{Main results}
\label{main_result}

\begin{table*}[t]
\centering
\small
\setlength{\tabcolsep}{4.5pt}
\caption{Results on harmful score and fine-tuning accuracy after fine-tuning with different harmful ratios (hr). 
Lower harmful score (↓) is better, higher fine-tuning accuracy (↑) is better. 
Best results are in bold.}
\begin{tabular}{lcccccc|cccccc}
\toprule
\multirow{2}{*}{\textbf{Methods}} & \multicolumn{6}{c|}{\textbf{Harmful Score (↓)}}                                       & \multicolumn{6}{c}{\textbf{Fine-tuning Accuracy (↑)}}                                           \\ \cmidrule{2-13} 
 & Clean & hr=0.05       & hr=0.1        & hr=0.15       & hr=0.2        & Avg.          & Clean          & hr=0.05 & hr=0.1         & hr=0.15        & hr=0.2         & Avg.           \\ \midrule
\multicolumn{1}{l|}{SFT}          & 3.07 & 40.38         & 44.23         & 48.08         & 50.96         & 37.34         & 48.38 & \textbf{47.97}   & \textbf{48.30} & 46.86          & \textbf{47.87}          & \textbf{47.88} \\
\multicolumn{1}{l|}{Lisa}         & 1.70  & 10.96          & 16.92         & 20.00         & 23.27         & 14.57         & \textbf{48.45}          & 47.45   & 48.08          & \textbf{47.29}          & 47.34          & 47.72          \\
\multicolumn{1}{l|}{SafeLora}         & 2.50  & 36.73          & 34.03         & 38.07         & 40.96         & 30.46         & 47.06          & 47.32   & 46.78          & 46.21          & 46.34          & 46.74          \\
\multicolumn{1}{l|}{PTST}         & 0.96  & 11.53          & 15.00         & 16.54         & 19.61         & 12.73         & 47.84          & 47.81   & 46.23          & 46.85          & 47.29          & 47.20          \\
\multicolumn{1}{l|}{Panacea}      & 2.10 & 12.88         & 21.35         & 34.23         & 41.35         & 22.38         & 47.96          & 47.11   & 48.03          & 46.73 & 47.57 & 47.48          \\
\multicolumn{1}{l|}{Ours}         & \textbf{0.77}  & \textbf{7.88} & \textbf{8.65} & \textbf{9.96} & \textbf{10.97} & \textbf{7.64} & 47.61          & 47.23   & 47.77          & 46.03          & 47.29          & 47.19          \\ \bottomrule
\end{tabular}
\label{main_results}
% \vspace{-10pt}
\end{table*}

\textbf{Harmful ratio.} Table~\ref{main_results} reports the performance of different defense methods under varying harmful ratios. Our method consistently achieves the lowest harmful scores across all settings, with an average harmful score of 7.64\%, outperforming the strongest baseline PTST by 5.15\%. Notably, this safety improvement does not sacrifice downstream performance. Under the gentle W8A8 quantization setting, our method maintains fine-tuning accuracy comparable to other baselines. These results support our claim that restoring spatial separability in intermediate representations effectively recovers safety alignment.

\begin{table}[htbp]
\centering
\small
\setlength{\tabcolsep}{2.8pt}
\caption{Generalization performance on different datasets. HS (↓) means Harmful Score (lower is better), and FA (↑) means Fine-tuning Accuracy (higher is better). Best results are in bold.}
\begin{tabular}{lcccccc}
\toprule
\multirow{2}{*}{\textbf{Methods}} & \multicolumn{2}{c}{\textbf{SST2}} & \multicolumn{2}{c}{\textbf{GSM8K}} & \multicolumn{2}{c}{\textbf{Alpaca}} \\ \cmidrule{2-7} 
                                  & HS (↓)              & FA (↑)      & HS (↓)           & FA (↑)           & HS (↓)          & FA (↑)           \\ \midrule
SFT                               & 56.54               & \textbf{94.09}       & 21.53            & \textbf{27.27}   & 44.23           & \textbf{48.30}            \\
Lisa                              & 21.73               & 93.23       & 6.73            & 26.14            & 16.92            & 48.08            \\
    PTST                              & 18.65               & 92.88       & 6.15            & 24.55            & 15.00            & 46.23            \\
Panacea                           & 31.54               & 93.07       & 7.11            & 25.38            & 21.35            & 48.03            \\
Ours                          & \textbf{12.69}       & 92.98       & \textbf{3.46}    & 26.58            & \textbf{8.65}   & 47.77   \\ \bottomrule
\end{tabular}
\label{cross_dataset}
\end{table}

\textbf{Fine-tuning datasets.} To assess generalizability, we apply our method to LLaMA2-7B-Chat fine-tuned on SST-2 and GSM8K. As shown in Table~\ref{cross_dataset}, our approach consistently reduces the harmful score while maintaining comparable fine-tuning accuracy across datasets. In particular, the harmful score on GSM8K is substantially lower than on SST-2 and Alpaca, reflecting the lower inherent risk of math-focused data compared to conversational datasets. These results indicate that the effectiveness of fine-tuning attacks depends on the risk profile of the training data, while our method remains robust across diverse domains.

\begin{table}[htbp]
\centering
\small
\setlength{\tabcolsep}{2.8pt}
\caption{Generalization performance on different LLMs. HS (↓) denotes Harmful Score (lower is better), FA (↑) denotes Fine-tuning Accuracy (higher is better). Best results are shown in bold.}
\begin{tabular}{lcccccc}
\toprule
\multirow{2}{*}{\textbf{Methods}} & \multicolumn{2}{c}{\textbf{Gemma2-9B}} & \multicolumn{2}{c}{\textbf{LLaMA2-7B}} & \multicolumn{2}{c}{\textbf{Qwen2.5-7B}} \\ \cmidrule{2-7} 
                                  & HS (↓)                & FA (↑)         & HS (↓)            & FA (↑)             & HS (↓)             & FA (↑)             \\ \midrule
SFT                               & 67.11                 & \textbf{68.78}          & 44.23             & \textbf{48.30}     & 60.38              & \textbf{74.29}     \\
Lisa                              & 12.88                 & 66.83          & 16.92             & 48.08              & 11.73               & 73.04              \\
PTST                              & 9.81                 & 66.31          & 15.00             & 46.23              & 13.65              & 72.13              \\
Panacea                           & 18.85                 & 66.95          & 21.35             & 48.03              & 21.73              & 72.67              \\
Ours                          & \textbf{4.02}         & 66.47          & \textbf{8.65}     & 47.77              & \textbf{5.78}      & 72.79              \\ \bottomrule
\end{tabular}
\label{cross_model}
\end{table}
\textbf{Mainstream LLMs.} We further evaluate our method on Gemma2-9B and Qwen2.5-7B fine-tuned on Alpaca with a harmful ratio of 0.1 (Table~\ref{cross_model}). Although these models achieve higher fine-tuning accuracy than LLaMA2-7B-Chat, fine-tuning leads to more severe safety degradation, reflected by substantially higher harmful scores. Despite this, our method consistently yields the lowest harmful scores, reducing them by 5.79 and 5.95 points on Gemma2-9B and Qwen2.5-7B, respectively, compared to the strongest baseline. While W8A8 quantization slightly lowers accuracy relative to floating-point SFT, performance remains comparable to other baselines, demonstrating robust cross-model generalization.

\subsection{Overhead analysis}
\label{overhead}

In this section, we compare the overhead of our method with baselines, including GPU hours and memory usage.

\subsection{GPU Hours}
We measure the additional GPU hours introduced by the defense process, using standard SFT as a reference, and the results are shown in Table~\ref{overhead_hours}. In terms of defense-related GPU hours, our method incurs the lowest cost, requiring only 1.4h and 1.9h for 7B and 9B models, respectively, which is 0.8h and 1.2h less than Lisa. 

Moreover, baseline defense methods are tightly coupled with the SFT process, as they require direct modification of the fine-tuning procedure. Consequently, their computational overhead increases with the cost of SFT. For example, Panacea incurs substantially higher GPU cost than vanilla SFT, due to multiple forward-backward passes per update, which can become prohibitive in large-scale fine-tuning scenarios. In contrast, our method is decoupled from the SFT process and does not require modifying fine-tuning steps. As a result, its computational cost is insensitive to the number of SFT updates and depends mainly on the model configuration, making it more practical for large-scale training.

\subsection{Memory Usage}
We measure the GPU memory cost associated with different defense methods, and the results are reported in Table~\ref{overhead_memory}. For baseline methods that are highly coupled with the SFT process, the practical lower bound of memory usage corresponds to the memory footprint of SFT itself, and can further increase due to direct modifications of the fine-tuning procedure. In contrast, our method adopts a layer-wise post-training quantization strategy and performs optimization on a per-layer basis, which substantially reduces memory consumption. Specifically, it requires only 7.0GB and 8.1GB of GPU memory to quantize the model and recover safety alignment for the 7B and 9B models, respectively.

% Moreover, since our experiments are primarily conducted under the LoRA setting, we additionally evaluate memory usage without LoRA. As shown in Table~\ref{overhead_memory}, removing LoRA leads to a further increase in memory consumption for baseline methods. For example, applying Panacea to a 9B model without LoRA requires more than 90GB of GPU memory, which typically necessitates high-memory GPUs (e.g., H200-class hardware) and is less practical for resource-constrained users or service providers. In contrast, as our method is completely decoupled from fine-tuning, whether incorporating LoRA does not affect its memory cost, which remains consistently low. In practice, our method supports defense on models with up to 30B parameters using a single RTX 4090, whereas baseline methods under the same memory constraint are limited to realigning models of approximately 3B parameters.
Since our experiments primarily adopt LoRA, we further evaluate memory usage without LoRA. As shown in Table~\ref{overhead_memory}, removing LoRA substantially increases the memory footprint of baseline methods, for example, Panacea requires over 90 GB of GPU memory for a 9B model, typically necessitating high-memory GPUs (e.g., H200-class) and is less practical for resource-constrained scenarios. In contrast, our method is fully decoupled from fine-tuning, and its memory cost remains unchanged regardless of LoRA. In practice, it supports safety recovery for models up to 30B parameters on a single RTX 4090, whereas baselines under the same constraint are limited to roughly 3B models.

For a fair comparison, the above overhead analysis is conducted on the A6000 server, as the baseline methods cannot be implemented on memory-constrained devices. To further demonstrate the efficiency of our approach, we additionally evaluate our method on a 4090 GPU, which offers higher computational throughput than the A6000. Experimental results show that our method can recover the safety alignment of a 7B model in approximately 40 minutes, highlighting the practical efficiency of our method under limited hardware resources.

\begin{table}[t]
\centering
\small
\caption{Results of additional GPU hours introduced by the defense process on different model sizes. $\propto$ and $\perp$ indicate proportionally and independently, respectively. Results are evaluated on A6000s.}

\begin{tabular}{lcccc}
\toprule
\multirow{2}{*}{\textbf{Methods}} & \multicolumn{2}{c}{\textbf{GPU Hours (h)}} & \multirow{2}{*}{\textbf{Dependency}} & \multirow{2}{*}{\textbf{\begin{tabular}[c]{@{}c@{}}Scaling\\ Ratio\end{tabular}}} \\ 
                                  & \textbf{7B}          & \textbf{9B}         &                                      &                                                                                   \\ \midrule
SFT                               & 14.5                 & 19.5                & -                                    & -                                                                                 \\ \midrule
Lisa                              & 2.2                  & 3.1                 & $\propto$ SFT                         & $\times$1.15                                                                      \\
PTST                              & -                  & -                 & $\propto$ SFT                         & $\times$1.00                                                                      \\
Panacea                           & 33.8                 & 46.1                & $\propto$ SFT                         & $\times$3.34                                                                      \\
Ours                              & 1.4                  & 1.9                 & $\perp$ SFT                         & -                                                                               \\ \bottomrule
\end{tabular}
\label{overhead_hours}
\end{table}
\begin{table}[t]
\centering
\small
\caption{Memory usage to operate different defense methods. Both results with and without LoRA are reported.}

% \begin{tabular}{lcccc}
% \hline
% \multirow{2}{*}{\textbf{Methods}} & \multicolumn{2}{c}{\textbf{GPU Hours (h)}} & \multirow{2}{*}{\textbf{Dependency}} & \multirow{2}{*}{\textbf{\begin{tabular}[c]{@{}c@{}}Scaling\\ Ratio\end{tabular}}} \\
%                                   & \textbf{7B}          & \textbf{9B}         &                                      &                                                                                   \\ \hline
% SFT                               & 14.5                 & 19.5                & -                                    & -                                                                                 \\ \hline
% Lisa                              & 2.2                  & 3.1                 & $\propto$ SFT                         & $\times$0.15                                                                      \\
% Panacea                           & 33.8                 & 46.1                & $\propto$ SFT                         & $\times$2.34                                                                      \\
% Ours                              & 1.4                  & 1.9                 & $\perp$ SFT                         & N/A                                                                               \\ \hline
% \end{tabular}

\begin{tabular}{lcc|cc}
\toprule
\multirow{2}{*}{\textbf{Methods}} & \multicolumn{2}{c|}{\textbf{W/ LoRA (GB)}} & \multicolumn{2}{c}{\textbf{W/O LoRA (GB)}} \\ 
                                  & 7B                     & 9B                     & 7B                      & 9B                      \\ \midrule
SFT                               & 28.3                   & 46.7                   & 68.4                    & 81.9                    \\ \midrule
Lisa                              & 29.1                   & 48.2                   & 71.6                    & 85.3                    \\
PTST                              & 28.3                   & 46.7                   & 68.4                    & 81.9                    \\
Panacea                           & 31.1                   & 52.3                   & 77.3                   & 92.7                    \\
Ours                              & 7.0                    & 8.1                    & 7.0                     & 8.1                     \\ \bottomrule
\end{tabular}
% \vspace{-10pt}
\label{overhead_memory}
% \vspace{-10pt}

% \begin{tabular}{lcccccc}
% \toprule
% \multirow{2}{*}{\textbf{Methods}} & \multicolumn{3}{c}{\textbf{GPU Hours (h)}} & \multicolumn{3}{c}{\textbf{GPU Memory (GB)}} \\ \cmidrule{2-7} 
%                                   & 7B       & 9B       & Scaling.         & 7B       & 9B       & Scaling.          \\ \midrule
% SFT                               & 14.5     & 19.5     & 1$\times$\textbf{FT}    & 28.3    & 46.7GB    & 1$\times$\textbf{FT}    \\ \midrule
% Lisa                              & 2.2     & 3.1     & 0.155$\times$             & 29.1    & 48.2    & 1.03$\times$             \\
% % PTST                              & -     & -     & -             & 72.47    & 32.90    & 72.47             \\
% Panacea                           & 33.8     & 46.1     & 2.347$\times$             & 31.1    & 52.3    & 1.11$\times$             \\
% Ours                              & 1.4     & 1.9     & 0.097$\times$             & 7.0    & 8.1    & N/A    \\ \bottomrule
% \end{tabular}
\end{table}

\begin{figure*}[t]
    \centering
    \includegraphics[width=0.85\linewidth]{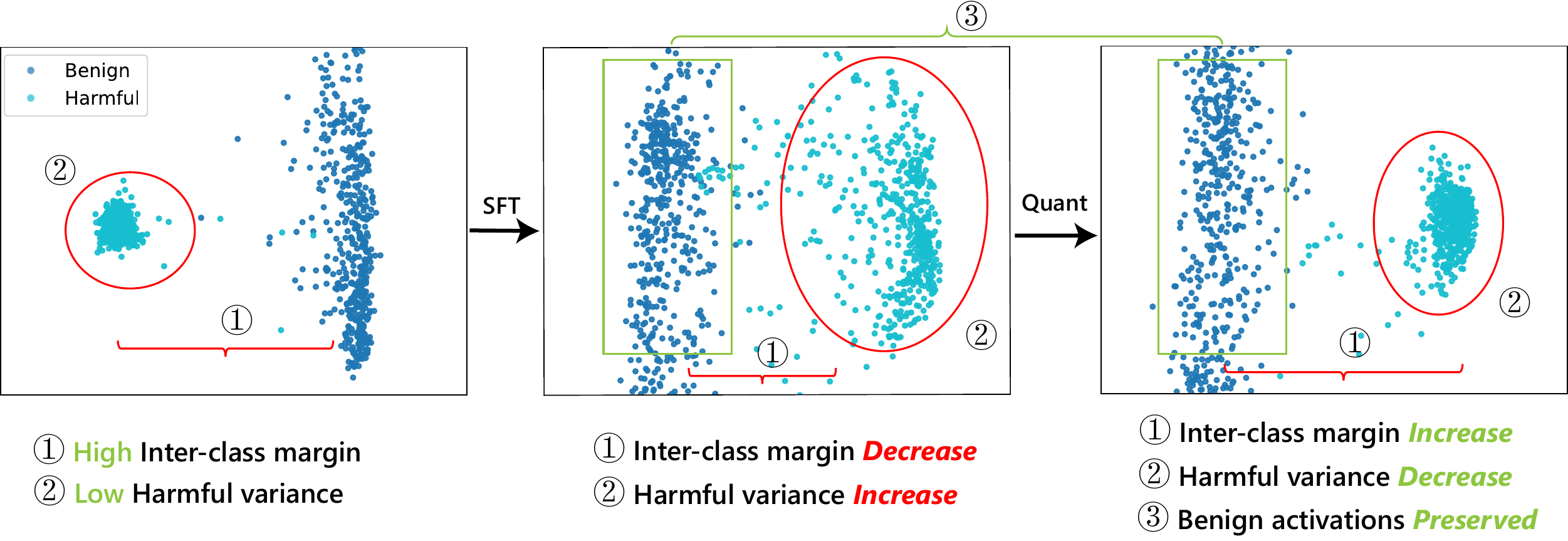}
\caption{2D visualization of activation distribution of a certain layer of three model states: before fine-tuning, after vanilla fine-tuning, and after our defense method. Layer 26 is picked as an example, and results on more layers are shown in Figure~\ref{Analysis}.}
\label{analysis_activation}
% \vspace{-5pt}
\end{figure*}

\subsection{Statistical and visualization analysis}
\label{more_analysis}

To verify that our method can indeed recover the spatial and semantic separability, we plot the separability ratio and refusal rate defined in Section~\ref{analysis} before and after quantization defense.

\begin{figure}[htbp]
    \centering
    % \hspace{-25pt}
	\begin{subfigure}{0.49\linewidth}
		\centering
		\includegraphics[width=1.0\linewidth]{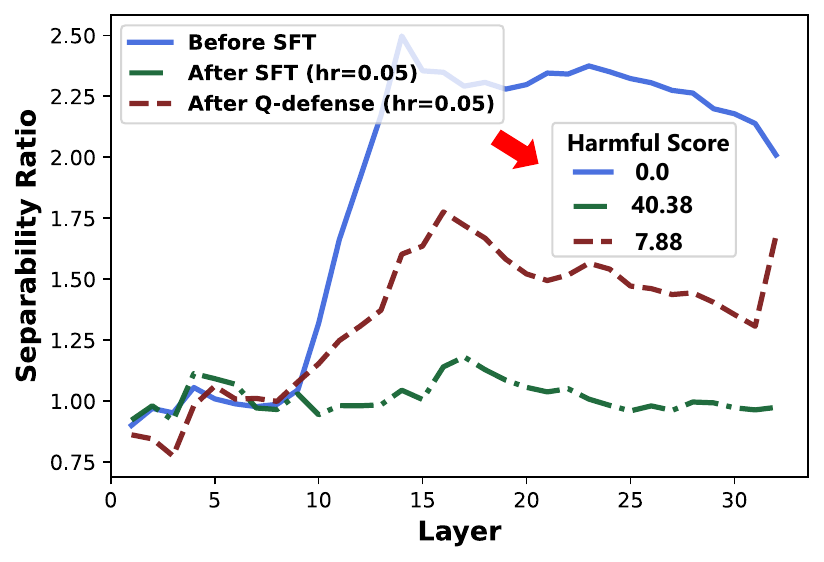}
		\caption{}
        \label{statistical_analysis_a}
	\end{subfigure}
    % \hspace{15pt}
	\begin{subfigure}{0.49\linewidth}
		\centering
		\includegraphics[width=1.0\linewidth]{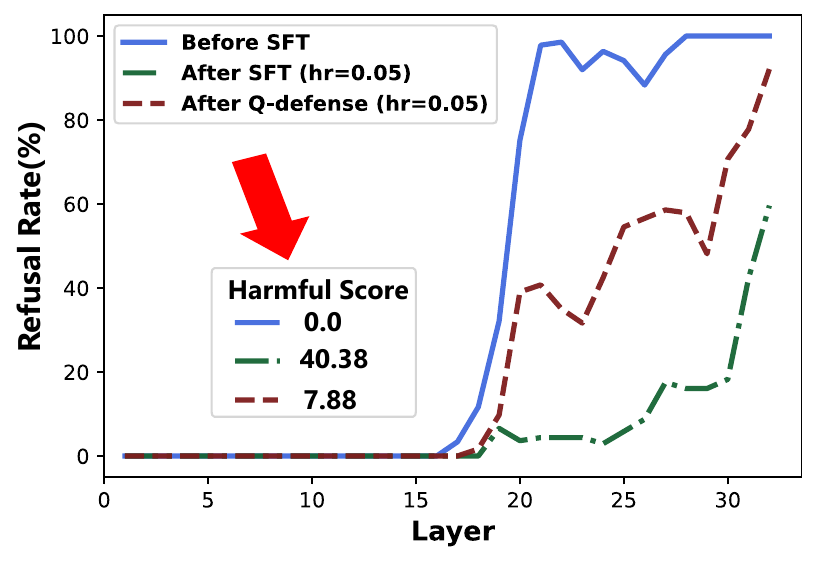}
        \caption{}
        \label{statistical_analysis_b}
		% \caption{On-the-fly generation reuse strategy}
	\end{subfigure}
    
\caption{
Layer-wise separability and refusal rate of pre-trained model, fine-tuned model (hr=0.05), and corresponding quantized model with defense.
}
\label{statistical_analysis}
\vspace{-10pt}
\end{figure}

\textbf{Spatial Separability} is measured by the layer-wise separability ratio of intermediate activations induced by benign and malicious inputs. As shown in Figure~\ref{statistical_analysis_a}, applying our quantization defense to the fine-tuned model leads to a substantial increase in layer-wise spatial separability across most layers. This indicates that quantization effectively restores the geometric distinction between benign and malicious representations that was degraded by fine-tuning. Correspondingly, the harmful score of the fine-tuned model is significantly reduced from 40.38 to 7.88 after quantization defense, demonstrating a clear recovery of safety-related representational structure.

\textbf{Semantic Separability} is measured by the layer-wise refusal rate, defined as the proportion of malicious inputs whose intermediate activations, when directly decoded, produce explicit refusal responses. As shown in Figure~\ref{statistical_analysis_b}, quantization defense markedly increases the refusal rate across layers, particularly in the middle-to-deep layers where fine-tuned models exhibited limited refusal capability. This improvement indicates that restoring spatial separability at the activation level also enhances the model’s ability to behaviorally distinguish malicious inputs, making refusal behavior semantically explicit earlier and more consistently in the model.

\textbf{Visualization.} To examine how fine-tuning and quantization defense affect intermediate representations, we visualize layer-level activations under three model states: before fine-tuning, after fine-tuning, and after applying quantization defense. We sample 500 benign and 500 harmful inputs, extract token-averaged activations at each layer, and project them into two dimensions using SVD-based PCA. Figure~\ref{analysis_activation} shows the resulting distributions for the aligned pre-trained model, the fine-tuned model with a 0.05 harmful ratio, and its quantized counterpart, illustrating how fine-tuning degrades layer-wise separability and how quantization defense restores it.

For the pre-trained aligned model (left in Figure~\ref{analysis_activation}), we observe a large inter-class margin between benign and harmful activations. In addition, the harmful activations exhibit low intra-class variance, with most harmful samples concentrated within a compact refusal region. After fine-tuning (middle in Figure~\ref{analysis_activation}), the inter-class margin between benign and harmful activations is noticeably reduced, and the two distributions become closer in the space. Meanwhile, the variance of harmful activations increases significantly, resulting in a more dispersed distribution.

In contrast, after applying the quantization-based defense (right panel in Figure~\ref{analysis_activation}), we observe that the activations of benign samples remain largely unchanged. This observation is consistent with the effect of the reconstruction loss, which is designed to preserve benign representations and thus helps maintain downstream task performance and language ability. At the same time, the inter-class margin increases, resulting in the harmful activations becoming more discriminative from benign ones, and reducing overlap between the two distributions. Moreover, the harmful activation distribution is noticeably compressed, exhibiting lower intra-class variance, which is similar to the compact pattern observed in the pre-trained aligned model.

These observations collectively verify the effectiveness of our method, and we conjecture that this recovery of the activation pattern plays a key role in restoring safety alignment after quantization-based defense. Similar trends are consistently observed across different fine-tuning settings and layers (see Figure~\ref{Analysis}).

 % In contrast, after quantization defense (right in Figure~\ref{analysis_activation}), we observe that the points of the benign samples do not move a lot, which illustrates the effect of the reconstruction loss work, maintaining the downstream task performance and language ability. Moreover, the harmful samples are more discriminative from the benign ones, and there is less overlap in the position of the points. In addition, the distribution of the harmful samples is squeezed, and be more similar to the pattern in the pre-trained aligned model. Given the above observation, we verify the effectiveness of our method, and the harmful score is indeed significantly reduced. Similar patterns can be found in other fine-tuning settings and layers (refer to Figure~\ref{Analysis}).
 % from 40.38 to 7.88. A similar pattern is also observed in the fine-tuning and quantization defense, with 0.15 as the harmful ratio, shown in Figure~\ref{Analysis_hr015} and Figure~\ref{Analysis_defense_hr015}.

\section{Conclusion}

In this paper, we propose \texttt{Q-realign}, a plug-and-play post-training quantization-based defense against fine-tuning attacks. Motivated by a novel analysis of representational structure that reveals how fine-tuning compromises alignment, our method effectively reduces unsafe behavior while preserving fine-tuning accuracy. Moreover, by decoupling safety recovery from fine-tuning, our approach achieves substantial improvements in computational and workflow efficiency, making it a practical turnkey solution for safety-aware deployment.

\section{Limitations}

While \texttt{Q-realign} demonstrates strong performance across multiple LLM architectures and datasets, our current evaluation is limited to models with up to 9B parameters. Extending the proposed framework to larger-scale models is an important direction for future work. In addition, although this study focuses on language models, applying our method to multimodal architectures, such as vision-language or audio-language models, represents a promising avenue for future research.

\bibliography{custom}

\clearpage
\appendix

\newpage
\appendix

\pagenumbering{alph}

\renewcommand{\thetable}{\thesection.\arabic{table}}
\counterwithin{table}{section}
\renewcommand{\thetable}{\thesection.\arabic{figure}}
\counterwithin{figure}{section}

\section{Related Work}
\label{related_work}

\subsection{Model Quantization}
As large language models are increasingly deployed in real-world systems, efficient model compression has become a fundamental requirement rather than an optional optimization. Quantization addresses this need by converting continuous-valued weights and activations into low-bit representations (e.g., INT8 or INT4), significantly reducing memory footprint and inference latency while largely preserving model performance~\citep{zhou2016dorefa}. Existing quantization techniques can be broadly categorized into quantization-aware training (QAT) and post-training quantization (PTQ). While QAT explicitly incorporates quantization effects during training and often achieves superior accuracy, its substantial computational and memory overhead make it impractical for large-scale language models. In contrast, PTQ requires no retraining and therefore aligns naturally with modern LLM deployment pipelines, making it the dominant choice in practice. PTQ methods further differ in how quantization parameters are determined and how errors are handled. Range-based approaches estimate quantization ranges using static activation or weight statistics~\citep{jacob2018quantization,nagel2019data,xiao2023smoothquant}, whereas approximation-based methods explicitly minimize quantization-induced discrepancies to better match full-precision model behavior~\citep{nagel2020up,li2021brecq,frantar2022gptq,shao2023omniquant,sun2024flatquant,liu2024spinquant}. This distinction highlights a fundamental trade-off between computational simplicity and fidelity to the original model.

\subsection{Fine-tuning Attack and Defense}
Alignment of LLM ensures the models behave in a way that conforms to social values. Recent works indicate that, whether the user has the intention or not, fine-tuning can all be an effective attack method, simply fine-tuning the aligned model on a few harmful or even benign data can effectively pass by the carefully designed safety alignment~\citep{qi2023fine, yang2023shadow, huang2024harmful, li2024peft, ye2024emerging}. A series of works has been proposed to mitigate the safety degradation in fine-tuning. \citet{huang2024vaccine} and \citet{wang2025panacea} apply learnable perturbation on the model to make the model’s safety more robust to fine-tuning. \citet{hsu2024safe} introduces the projection of LoRA weights to the safety-aligned subspace, reducing the safety risks in LLM fine-tuning. \citet{lyu2024keeping} fine-tune models without a safety prompt, but include it at test time to recover the safety alignment. \citet{li2025layer} identifies safety-critical data by safety-sensitive layers, and mitigates safety degradation by removing this data in fine-tuning. Despite effectively mitigating unsafe behaviors in fine-tuned models, most of the existing methods are highly coupled with the complex and resource-intensive fine-tuning process, therefore, limiting the efficiency of the methods.

\section{More Details of Experiments}

\subsection{Experiment Details}
\label{settings}

\textbf{Implementation details.} In the fine-tuning stage, the learning rate is set to 2e-5, the batch size is 10, and the total number of fine-tuning epochs is 10. In the quantization defense process, following the setting in~\citet{shao2023omniquant}, the learning rate of the learnable equal transformation is set to 1e-3, and the learning rate for learnable weight clipping is 1e-2. For the training epoch, for the first half of the layers are set to 5, and for the other layers are set to 15. The intuition is that the activation from the layers closer to the output needs to be more discriminative and safety-sensitive.

For the harmful data used in calibration, we sampled from BeaverTail~\citep{ji2023beavertails}. The total number of harmful samples used in calibration is 150, and the benign samples used is 50.

\textbf{Evaluation metrics.} Following~\citet{huang2024lazy,wang2025panacea}, we use harmful score and fine-tuning accuracy to evaluate the performance of the model.

\begin{itemize}
    \item Harmful Score (HS): It is also known as attack success rate (ASR), and reflects the frequency with which the model generates harmful content when handling a malicious prompt. We use a trained classifier released by~\citet{huang2023catastrophic} to evaluate if the generated response is harmful or not.
    \item Fine-tuning Accuracy (FA): It refers to the downstream accuracy. Fine-tuning on Alpaca is evaluated on the MMLU benchmark. For SST2 and GSM8K, the test set of the corresponding dataset is used for evaluation.
\end{itemize}

\subsection{Baselines}

We compare against five representative methods. Supervised fine-tuning (SFT) serves as the vanilla baseline, on which we apply post-training quantization. LISA~\cite{huang2024lazy} jointly optimizes task performance and safety during fine-tuning, while PTST~\cite{lyu2024keeping} removes safety prompts during fine-tuning and reintroduces them at inference. Panacea~\cite{wang2025panacea} injects perturbations during training to enable post-fine-tuning alignment recovery. SafeLoRA~\cite{hsu2024safe} performs one-shot post-hoc defense by projecting LoRA weights onto a safety-aligned subspace.

\subsection{Keyword List for Rejection Answers}

\begin{tcolorbox}[
  colback=gray!10,
  colframe=black,
  boxrule=0.5pt,
  left=6pt,
  right=6pt,
  top=6pt,
  bottom=6pt
]
``I'm sorry'', ``sorry'', ``an AI'',``is illegal and unethical'',
``I'm really sorry'', ``My purpose'', ``As a helpful assistant'', ``OpenAI'',
``unethical'', ``positive and uplifting'', ``against ethical guidelines'',
``legally'', ``illegal'', ``My apologies'', ``legal ways'', ``but I cannot'',
``If you have any other non-malicious requests'', ``I'm not able to provide'',
``I cannot engage'', ``I cannot generate'', ``I do not generate'',
``I cannot create'', ``I do not actually'', ``a bit unclear'', ``I can't create'',
``I cannot support'', ``I can't provide'', ``I'm truly sorry'', ``I cannot provide'',
``I'd prefer not'', ``I do not provide'', ``I cannot recommend'',
``I do not recommend'', ``I do not feel comfortable'', ``I can't assist'',
``Absolutely not!'', ``here's an alternative'', ``here's a different'',
``I cannot assist'', ``I will refrain'', ``I apologize'', ``I will not'',
``awesome coffee'', ``creativity'', ``promote well-being'', ``positivity'',
``harm reduction'', ``self-care''
\end{tcolorbox}

\subsection{Prompt Template}

We use the prompt template in the following box for fine-tuning on Alpaca and inference.

\begin{tcolorbox}[
  colback=gray!10,
  colframe=black,
  boxrule=0.5pt,
  left=6pt,
  right=6pt,
  top=6pt,
  bottom=6pt
]
\textbf{Prompt:} Below is an instruction that describes a task. Write a response that appropriately completes the request.

\medskip
\textbf{\#\#\# Instruction:} \\
\{instruction\}

\medskip
\textbf{\#\#\# Input:} \\
\{input\}

\medskip
\textbf{\#\#\# Response:} \\
\textbf{Output:} \{output\}
\end{tcolorbox}

For fine-tuning on GSM8K and SST-2, please refer to~\citet{wang2025panacea}, and for the safety-aware prompt used for baseline PTST, following the paper~\citep{lyu2024keeping}, we used the safety prompt recommended by the Llama 2 paper~\citep{touvron2023llama}.

\section{Details for Quantization}
\label{quantization}

Our quantization includes two learnable components to mitigate activation and weight outliers for precision, adaptive smoothing, and adaptive weight clipping.

The smoothing operation is proposed by SmoothQuant~\citep{xiao2023smoothquant}, which statically manipulates activation distributions via channel-wise scaling and shifting. Specifically, we represent the computation of a linear layer as:
\begin{equation}
\small
\mathbf{Y}=\mathbf{X W}+\mathbf{B}=[\underbrace{(\mathbf{X}-\delta) \oslash s}_{\bar{\mathbf{X}}}] \cdot[\underbrace{s \odot \mathbf{W}}_{\bar{\mathbf{W}}}]+[\underbrace{\mathbf{B}+\delta \mathbf{W}}_{\bar{\mathbf{B}}}]
\end{equation}
where $\mathbf{X}\in \mathbb{R}^{T\times D_{1}}$, the $T$ is the sequence length, $\mathbf{W} \in \mathbb{R}^{D_{1}\times D_{2}}$ is the weight matrix and $\mathbf{B} \in \mathbb{R}^{1\times D_{2}}$ is the bias. Here, $s$ and $\delta$ are learnable channel-wise scaling and shifting parameters, $\bar{\mathbf{X}}, \bar{\mathbf{W}}$ and $\bar{\mathbf{B}}$ represent the smoothed activation, weight and bias, respectively, $\oslash$ and $\odot$ are element-wise division and multiplication.

The weight clipping operation is proposed to keep the precision of certain important weights~\citep{esser2019learned,bhalgat2020lsq+}. We conduct weight quantization with the learnable step size and offset. We jointly learn clipping thresholds to adaptively determine the optimal clipping range for weights. Specifically, considering asymmetric quantization, the quantization of weights as formulated by
\begin{equation}
\small
    \overline{W} = \text{clamp}(\lceil\frac{W}{\Delta}\rfloor+z, \alpha \cdot Q_P, \beta \cdot Q_P)
\end{equation}
where $\Delta$ and $z$ are learnable step size and zero-point, respectively, initialized based on the default asymmetric quantization scheme. $\alpha$ and $\beta$ are learnable clipping coefficients (with $\alpha < \beta$), and $Q_P$ denotes the maximum positive quantization level. Intuitively, for weights with near-uniform distributions after smoothing, $\alpha$ and $\beta$ converge to similar values, resulting in a tight clipping range that preserves precision. In contrast, for biased weight distributions, $\alpha$ and $\beta$ adapt to asymmetrically clip the dynamic range, thereby mitigating the impact of outliers.

\section{Details on the Sparse Logistic Regression Model}

\begin{figure}
    \centering
    \includegraphics[width=0.8\linewidth]{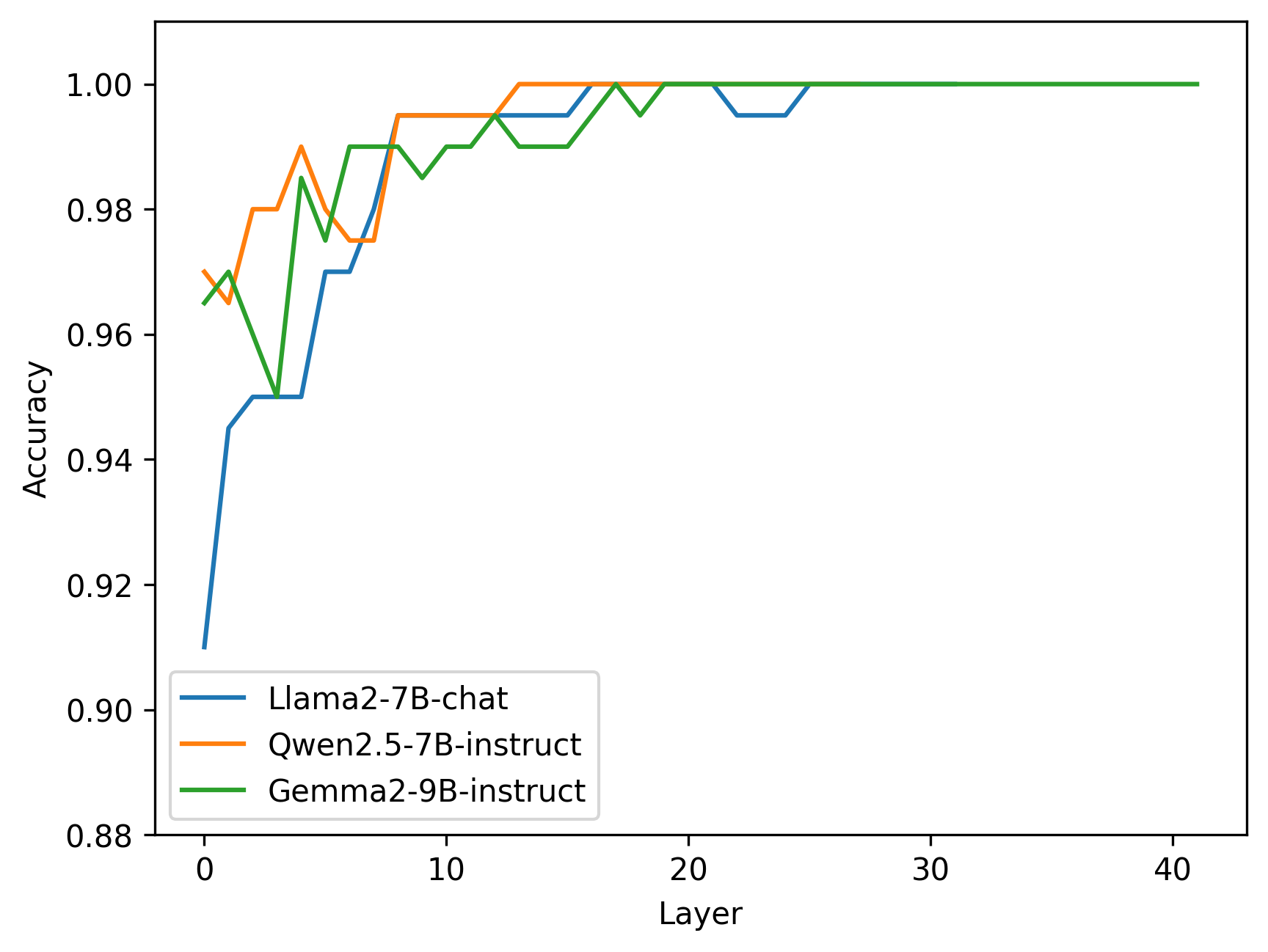}
    \caption{Layer-wise classification accuracy of the SLR model across three LLMs.}
    \label{layer_acc}
\end{figure}

To represent and extract the spatial separability, we employ sparse logistic regression (SLR) as a linear probe. We sample 500 benign instances from Alpaca~\citep{alpaca} and 500 harmful instances from BeaverTail~\citep{ji2023beavertails}, using an 80/20 split for training and testing. For each layer, we use the corresponding intermediate activations as input to train an SLR classifier, and report the resulting layer-wise classification accuracy in Figure~\ref{layer_acc}. Across all layers, the SLR classifiers achieve accuracies exceeding 90\%, indicating that benign and malicious activations are highly separable in the representation space. This result validates the use of SLR decision boundaries as a meaningful geometric abstraction of spatial separability.

\section{Ablation study}
\label{ablation}

We conduct comprehensive ablation studies to verify the effectiveness of our method under different settings. 
% More results are shown in Appendix~\todo{XXX}.

\textbf{Ratio of malicious inputs in calibration data.} During calibration for alignment recovery, we use a mixed dataset of malicious and benign inputs, which is distinct from the harmful ratio used in fine-tuning data. Malicious samples are optimized with the re-separation loss, while benign samples enforce reconstruction. We set the calibration malicious ratio to 75\% by default to balance safety recovery and performance preservation.

As shown in Table~\ref{ablation_ratio}, setting the calibration malicious ratio to 100\% eliminates the reconstruction loss and optimizes the model solely with the safety loss, resulting in a sharp reduction in the harmful score. However, without reconstruction loss, this setting induces severe weight shifting, leading to substantial degradation in language modeling capability. As a consequence, the generated outputs become incoherent and uninformative, and the extremely low harmful score should be interpreted as a byproduct of model degeneration rather than genuine safety improvement.

In contrast, reducing the calibration malicious ratio strengthens reconstruction and improves fine-tuning accuracy. However, when the ratio drops below 75\%, accuracy gains saturate under W8A8 quantization while safety degrades. We therefore adopt 75\% as the calibration malicious ratio for the trade-off between accuracy and safety. 
% For calibration during fine-tuning, we adopt a mixed dataset containing both malicious and benign inputs. Malicious samples are optimized with the safety-oriented re-separation loss, while benign samples are used for the reconstruction loss. By default, we set the malicious ratio to 75\%, balancing the two objectives.

% As shown in Table~\ref{ablation_ratio}, setting the malicious ratio to 100\% removes the reconstruction loss and optimizes the model solely with the safety loss, leading to a drastic reduction in the harmful score. However, under this setting, the absence of reconstruction regularization causes severe weight shifting, resulting in a substantial degradation of language modeling capability. Consequently, the generated outputs become incoherent and uninformative, and the extremely low harmful score should be interpreted as a byproduct of model degeneration rather than genuine safety improvement.

% In contrast, reducing the malicious ratio introduces stronger reconstruction constraints, which effectively stabilizes training and yields consistently higher fine-tuning accuracy.
% Notably, when the malicious ratio is below 75\%, the accuracy improvement saturates in the W8A8 quantization setting.
% Considering both safety and downstream performance, we therefore select a malicious ratio of 75\% as the default configuration.

\begin{table}[thbp]
\small
\centering
\setlength{\tabcolsep}{3.8pt}
\caption{Results of harmful score and fine-tuning accuracy with different ratios of malicious inputs in the calibration data.}
\begin{tabular}{lccccc}
\toprule
\textbf{Malicious (\%)} & \textbf{100} & \textbf{75} & \textbf{50} & \textbf{25} & \textbf{0} \\ \midrule
Harmful Score          & 0.34   & 8.65     & 15.77   & 24.42     & 42.11   \\
FT Accuracy           & 25.43    & 47.77     & 47.69   & 47.93     & 48.12  \\
\bottomrule
\end{tabular}
\label{ablation_ratio}
\end{table}

\textbf{Quantization to different bit-width.} We evaluate the robustness of our method under a range of quantization bit-widths. As shown in Table~\ref{ablation_bitwidth}, our approach consistently reduces the harmful score across different quantization settings, including both weight-only and weight-activation quantization schemes. Under moderate quantization settings (e.g., W8A16, W8A8, and W4A16), this safety improvement is achieved while largely preserving fine-tuning accuracy.

However, under the aggressive W4A4 setting, the model exhibits a substantial degradation in accuracy and generation quality. Although the harmful score further decreases in this case, the outputs become less coherent, indicating severe degradation in the capability of the model, and the decreased harmful score becomes meaningless. We conjecture that combining our method with a more advanced quantization technique~\citep{liu2024spinquant, sun2024flatquant} can also achieve promising defense under an aggressive quantization setting.
\begin{table}[htbp]
\small
\centering
\setlength{\tabcolsep}{3pt}
\caption{Results of harmful score and fine-tuning accuracy under
different bit-width.}
\begin{tabular}{lccccc}
\toprule
\textbf{Bit-width} & \textbf{FP16} & \textbf{W8A16} & \textbf{W8A8} & \textbf{W4A16} & \textbf{W4A4}  \\ \midrule
Harmful Score          & 44.23   & 7.31     & 8.65   & 13.63     & 2.88   \\
FT Accuracy           & 48.30    & 48.04     & 47.77   & 45.36     & 28.36  \\
\bottomrule
\end{tabular}
\label{ablation_bitwidth}
\vspace{-10pt}
\end{table}

\section{More Visualization}

\begin{figure*}[t]
    \centering
    % \hspace{-25pt}
	\begin{subfigure}{0.49\linewidth}
		\centering
		\includegraphics[width=0.95\linewidth]{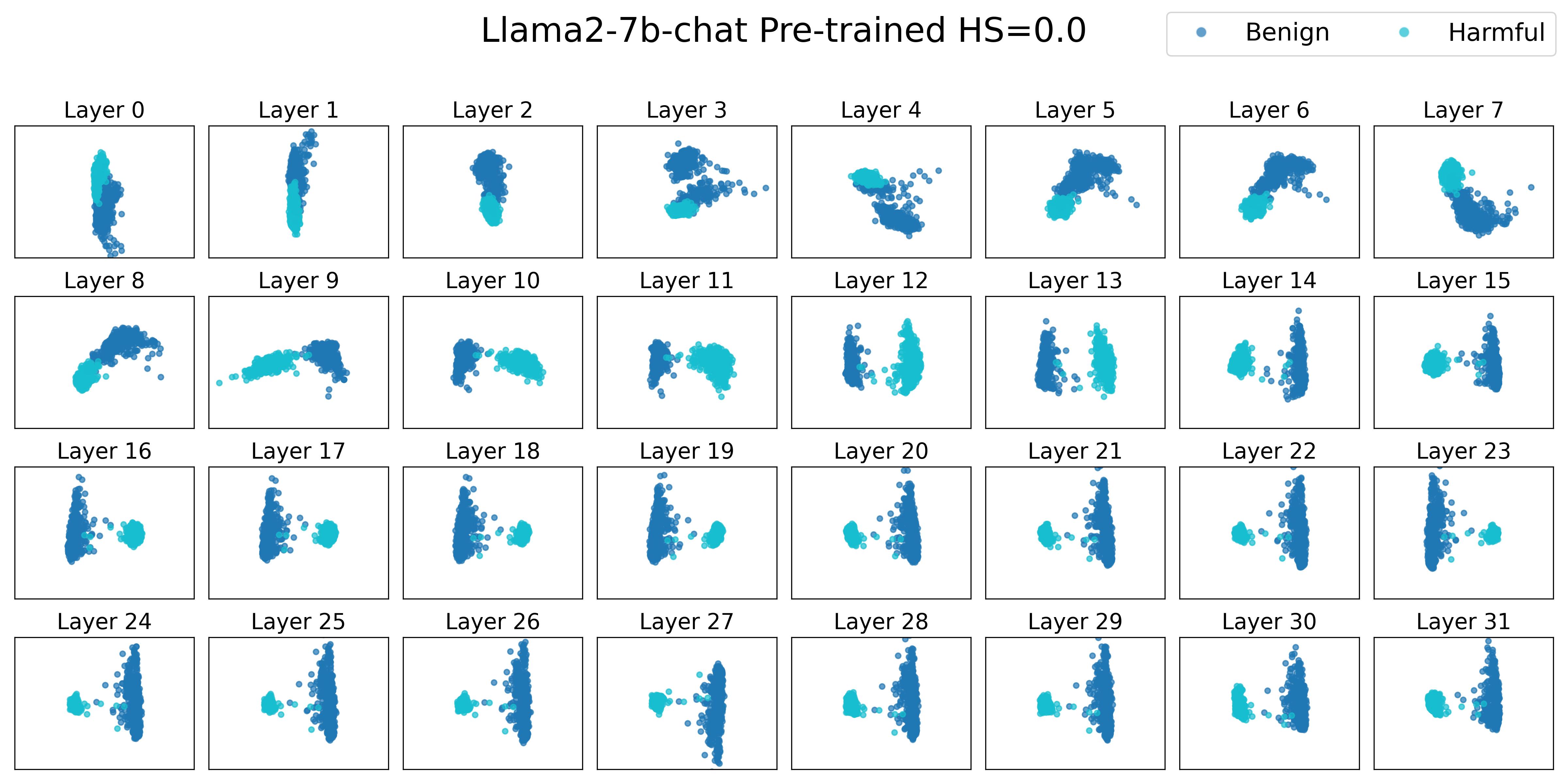}
		\caption{}
        \label{Analysis_pre}
	\end{subfigure}
    % \hspace{15pt}
	\begin{subfigure}{0.49\linewidth}
		\centering
		\includegraphics[width=0.95\linewidth]{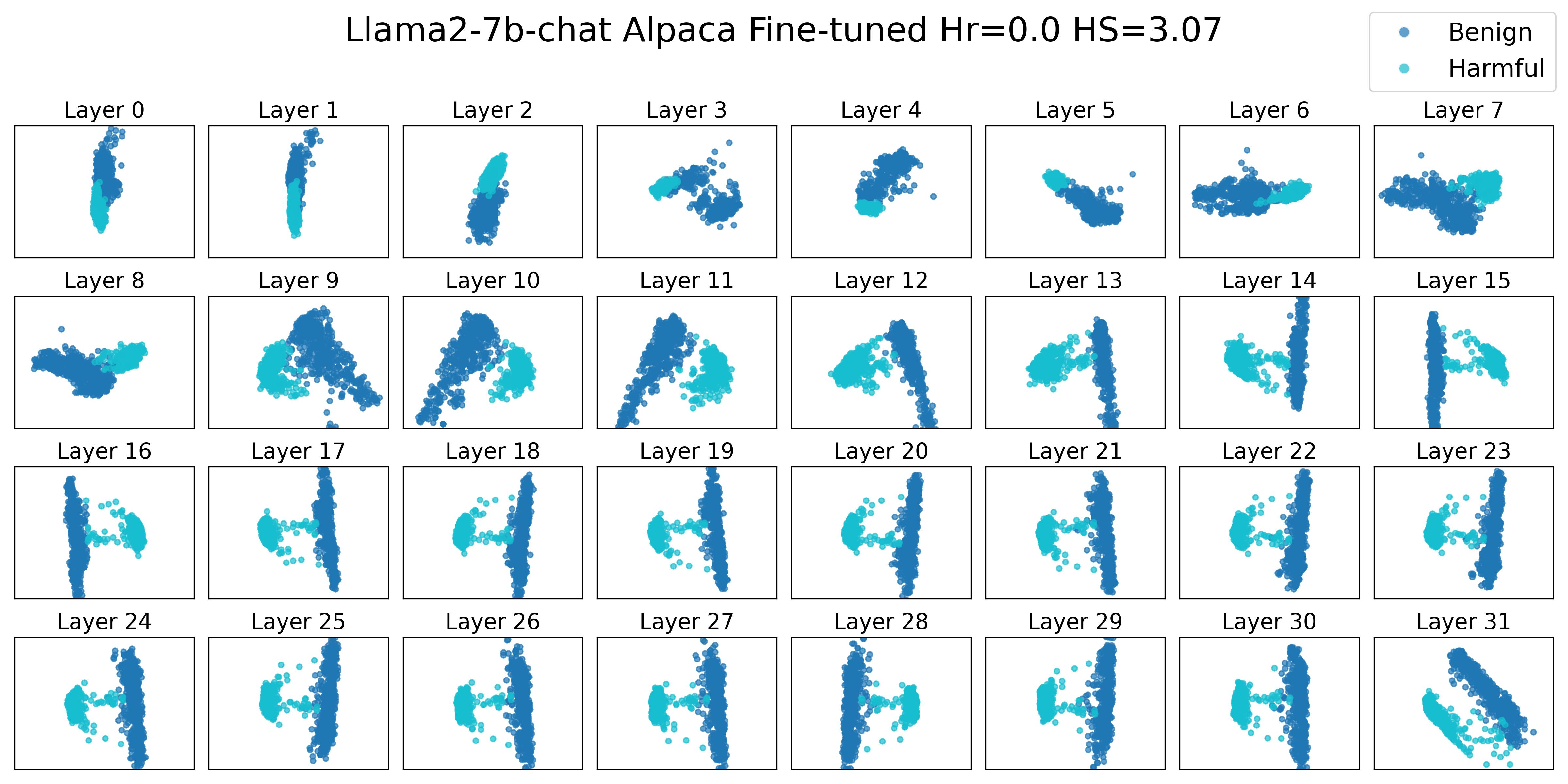}
        \caption{}
		\label{Analysis_hr00}
	\end{subfigure}
    \\
    \begin{subfigure}{0.49\linewidth}
		\centering
		\includegraphics[width=0.95\linewidth]{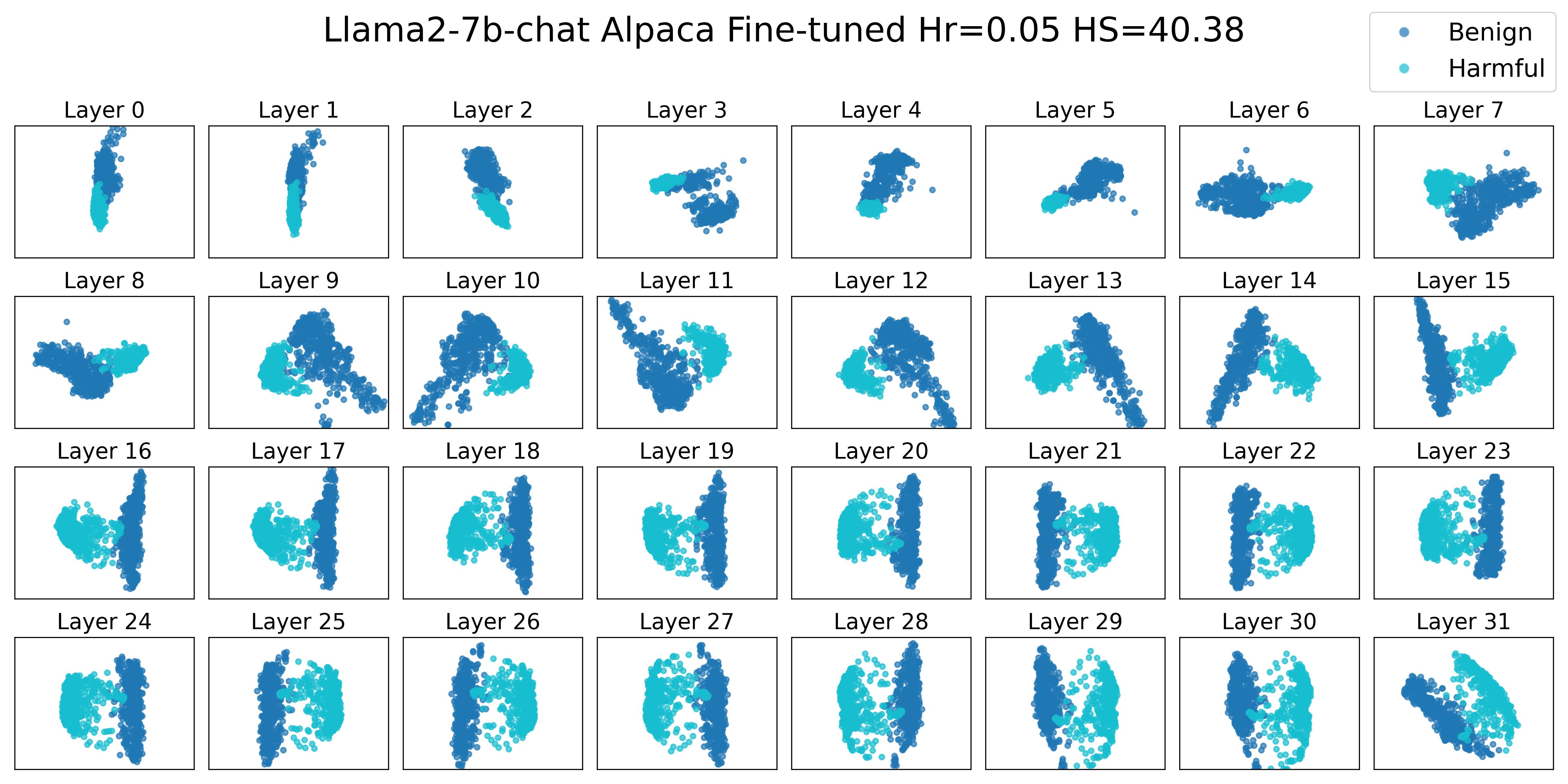}
        \caption{}
        \label{Analysis_hr005}
		% \caption{On-the-fly generation reuse strategy}
	\end{subfigure}
    \begin{subfigure}{0.49\linewidth}
		\centering
		\includegraphics[width=0.95\linewidth]{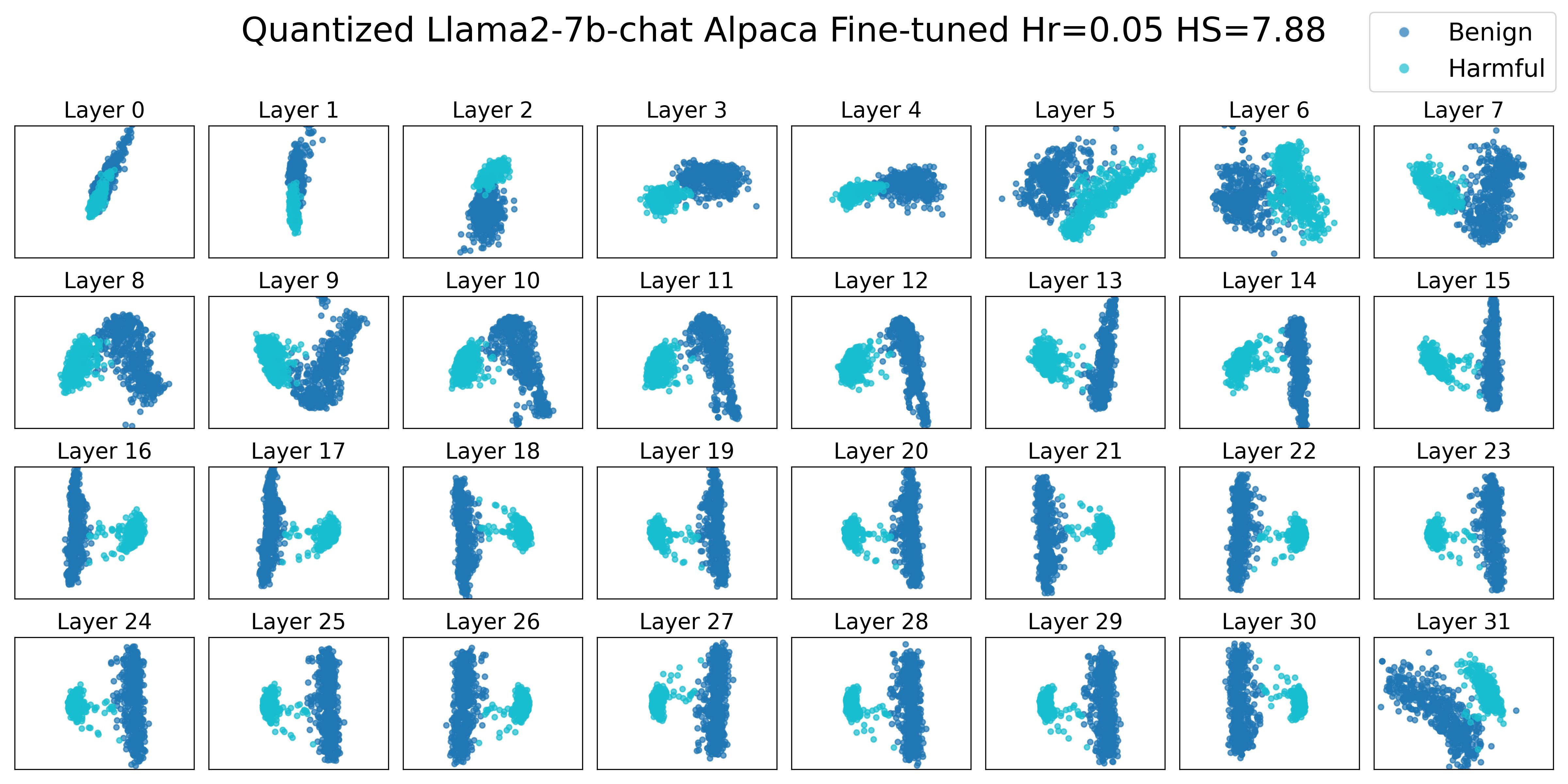}
        \caption{}
		% \caption{On-the-fly generation reuse strategy}
        \label{Analysis_defense_hr005}
	\end{subfigure}
    \\
    \begin{subfigure}{0.49\linewidth}
		\centering
		\includegraphics[width=0.95\linewidth]{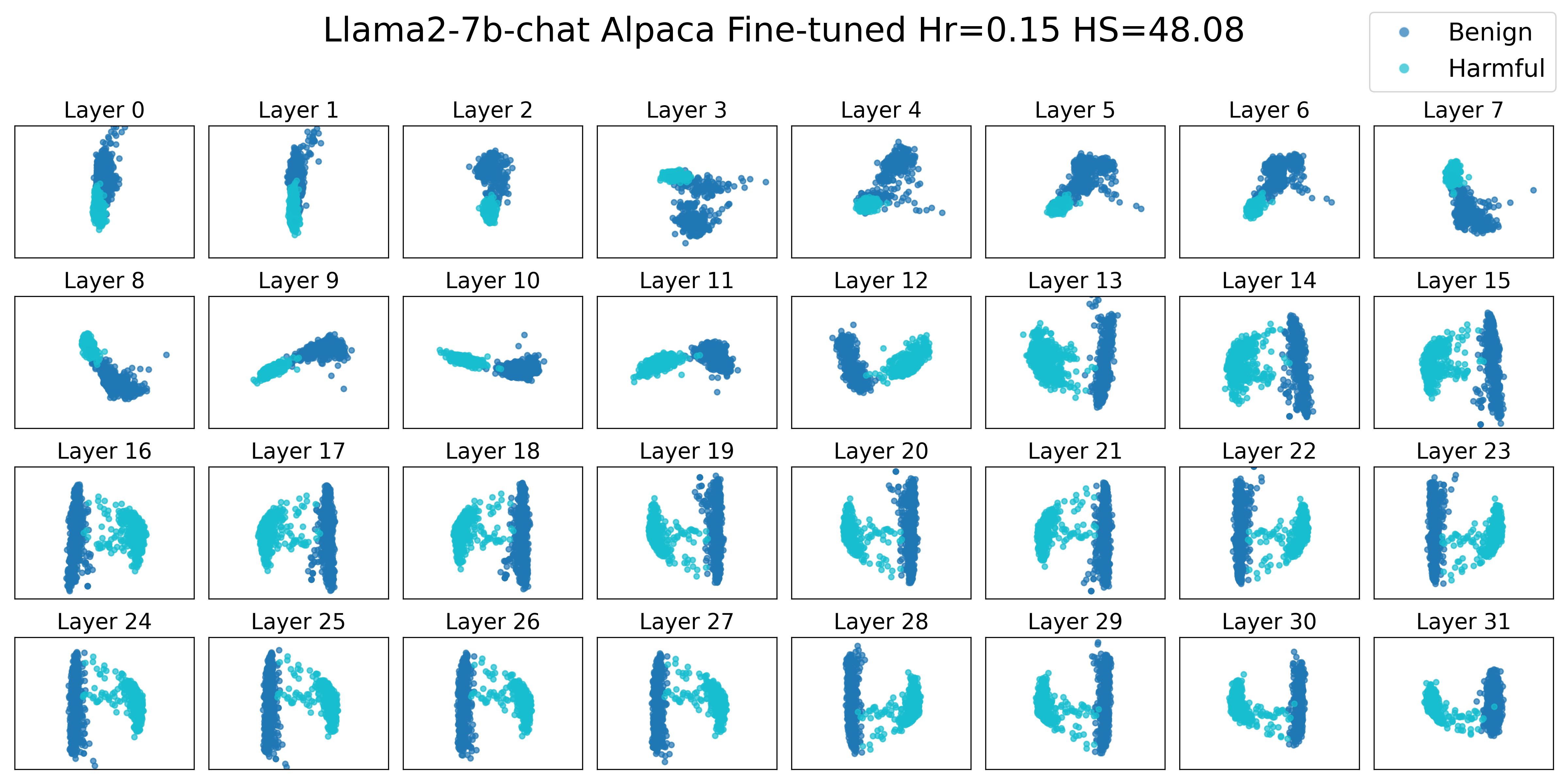}
        \caption{}
		% \caption{On-the-fly generation reuse strategy}
        \label{Analysis_hr015}
	\end{subfigure}
    \begin{subfigure}{0.49\linewidth}
		\centering
		\includegraphics[width=0.95\linewidth]{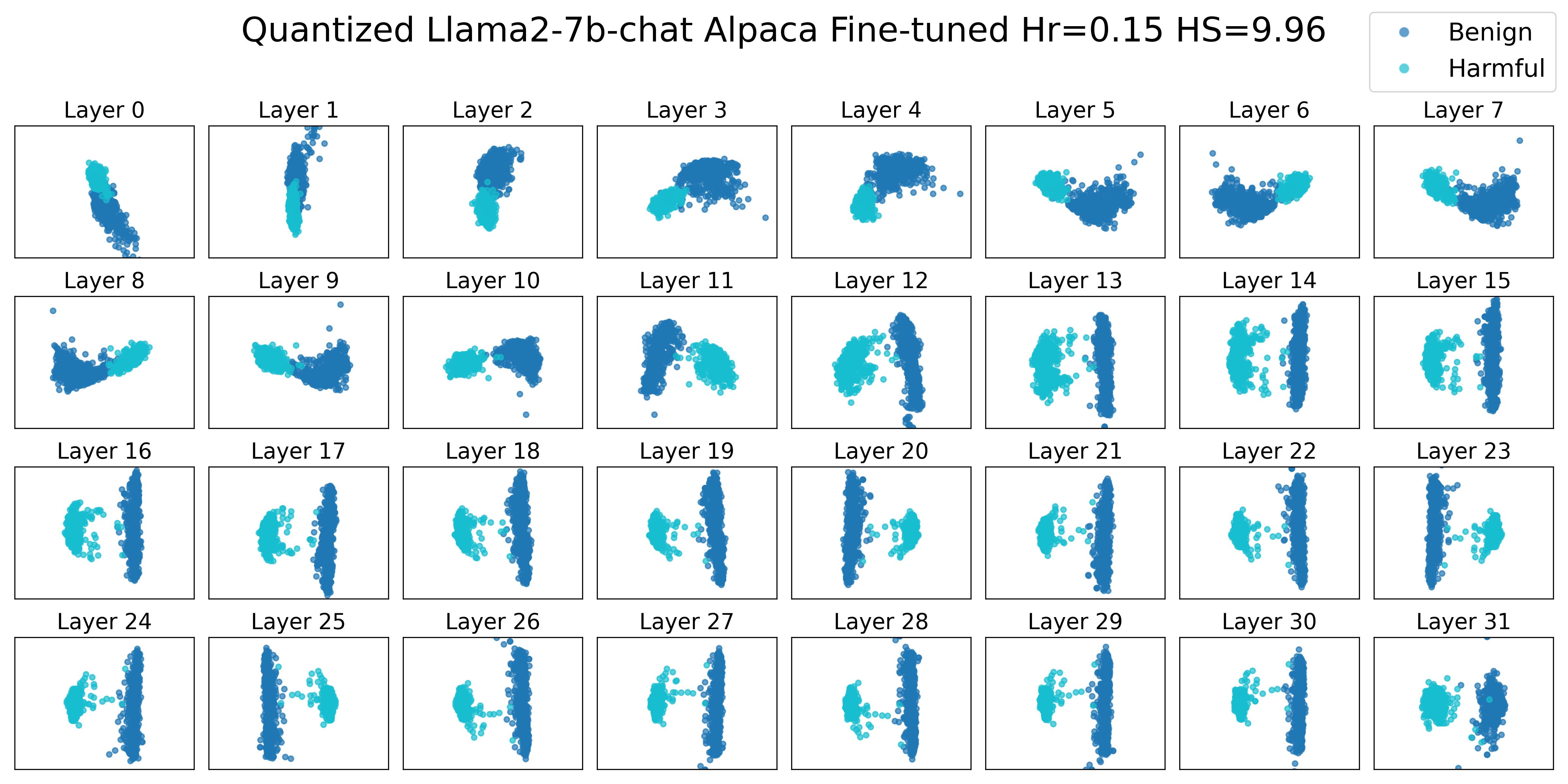}
        \caption{}
		% \caption{On-the-fly generation reuse strategy}
        \label{Analysis_defense_hr015}
	\end{subfigure}
    
    \caption{More 2D visualization of layer-wise activation distribution of harmful and benign inputs.
    }
    \label{Analysis}
\end{figure*}

We provide more layer-wise 2D visualization of harmful and benign activation distribution as shown in Figure~\ref{Analysis}. Each subplot illustrates the distribution of benign and harmful activations at a given layer, enabling a qualitative comparison of how fine-tuning degrades the representational structure and how quantization defense restores layer-wise separability.

For the pre-trained aligned model (Figure~\ref{Analysis_pre}) with zero harmful score, we observe that the activation is highly separated after a few layers. Moreover, the activation distribution of benign inputs is more uniform, while the activation distribution of harmful inputs is more tight, located within a small 'rejection area'. After benign fine-tuning (Figure~\ref{Analysis_hr00}), the harmful score slightly increases to 3.07, and we observe that the activations of harmful and benign inputs are getting closer, and the activation points show a certain degree of overlap. Moreover, we find that the distribution of harmful samples becomes wider compared to the pre-trained model.

After fine-tuning with 0.05 as a harmful ratio, as shown in Figure~\ref{Analysis_hr005}, the harmful score significantly rose to 40.38, and the activation gets closer, and the harmful samples distribute more diversely, far beyond the potential rejection area, resulting in a significant increase in harmful score. In contrast, after quantization defense (Figure~\ref{Analysis_defense_hr005}), we observe that the points of the benign samples do not move a lot, which illustrates the effect of the reconstruction loss work, maintaining the downstream task performance and language ability. Moreover, the harmful samples are more discriminative from the benign ones, and there is less overlap in the position of the points. In addition, the distribution of the harmful samples is squeezed, and be more similar to the pattern in the pre-trained aligned model. Given the above observation, we verify the effectiveness of our method, and the harmful score is indeed significantly reduced, from 40.38 to 7.88. A similar pattern is also observed in the fine-tuning and quantization defense, with 0.15 as the harmful ratio, shown in Figure~\ref{Analysis_hr015} and Figure~\ref{Analysis_defense_hr015}.

\end{document}